\documentclass[10pt]{article}
\usepackage[utf8x]{inputenc}
\usepackage{amsmath}
\usepackage{mathtools}
\usepackage{graphicx}
\usepackage{titling}
\graphicspath{{images/}}
\usepackage{parskip}
\usepackage{fancyhdr}
\usepackage{vmargin}
\usepackage{amsmath}
\usepackage{cite}      
\usepackage[table,xcdraw]{xcolor}
\usepackage{psfrag}    
\usepackage{mwe} 
\usepackage{caption}
\usepackage{bm} 
\usepackage{mathtools}
\usepackage{ amssymb }
\usepackage{stfloats} 
\usepackage{float}
\usepackage{placeins}
\usepackage{pdfpages}
\usepackage{multicol,listings}
\usepackage[english]{babel}
\usepackage{subfigure}
\setmarginsrb{1.75 cm}{1.25 cm}{1.75 cm}{1.25 cm}{0.5 cm}{1 cm}{0.5 cm}{1.5 cm}
\PassOptionsToPackage{english}{babel}
\title{ Topology of Syntax Networks across Languages}								

\makeatletter 
\let\thetitle\@title
\let\theauthor\@author
\let\thedate\@date
\makeatother

\begin{document}

\begin{titlepage}
	\centering
    \vspace*{0.5 cm}
    \includegraphics[scale = 0.5]{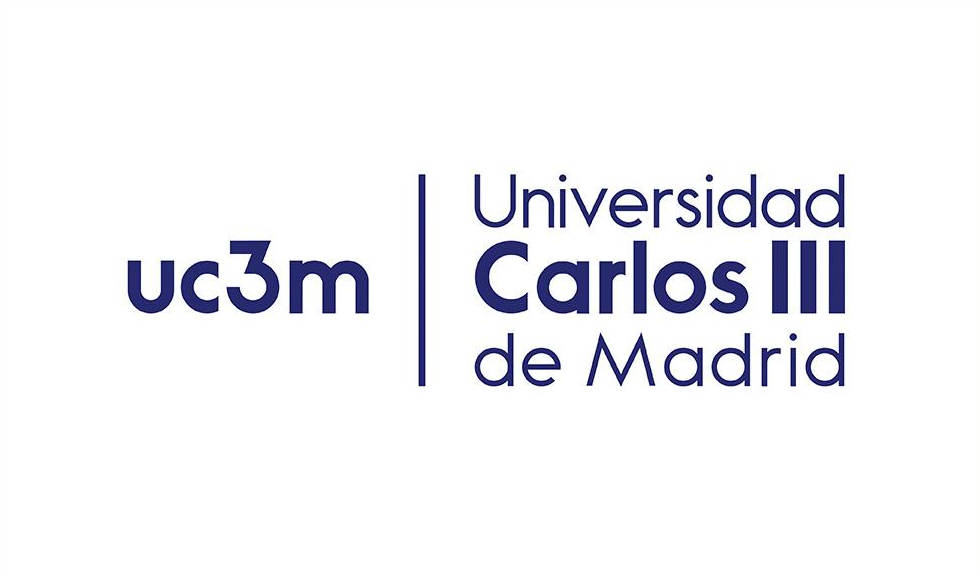}\\[1.0 cm]	
    \textsc{\LARGE Universidad Carlos III de Madrid}\\[2.0 cm]	
	\textsc{\Large Trabajo de fin de Máster}\\[0.5 cm]				
	\textsc{\large Masters Degree in Computational and Applied Mathematics}\\[0.5 cm]				
	\rule{\linewidth}{0.2 mm} \\[0.4 cm]
	{ \huge \bfseries \thetitle}\\ \vspace{0.5cm}
	\rule{\linewidth}{0.2 mm} \\[1.6 cm]

	\centering
	\emph{\textbf{Author:}}\\ \vspace{0.5cm}
	 \textbf{Juan Soria Postigo} \\
	\vspace{0.5cm}
	\emph{\textbf{Supervisors:}}\\ \vspace{0.5cm}
	 \textbf{Luís F. Seoane} \\
	 \textbf{José Cuesta}

	\vfill
	
\end{titlepage}

\thispagestyle{plain}
\begin{center}
    
    \large
    \vspace{0.9cm}
    \textbf{Acknowledgements}
\end{center}
Mis más sinceros agradecimientos a Luíño por su constante apoyo y contagiante pasión por su trabajo. Ha sido un placer hacer la tesis contigo.

Por otro lado agradecer a la UC3M y al banco Santander por la concesión de ayuda al estudio. Sin este soporte económico no me habría sido posible cursar el máster y estaré eternamente agradecido por ello.

\pagebreak

\thispagestyle{plain}
\begin{center}
    
    \large
    \vspace{0.9cm}
    \textbf{Abstract}
\end{center}
Syntax connects words to each other in very specific ways. Two words are syntactically connected if they depend
directly on each other. Syntactic connections usually happen within a sentence. Gathering all those connection across several sentences gives birth to syntax networks. Earlier studies in the field have analysed the structure and properties of syntax networks trying to find clusters/phylogenies of languages that share similar network features. The results obtained in those studies will be put to test in this thesis by increasing both the number of languages and the number of properties considered in the analysis. Besides that, language networks of particular languages will be inspected in depth by means of a novel network analysis\cite{seoane2022tcs}. Words (nodes of the network) will be clustered into topological communities whose members share similar features. The properties of each of these communities will be thoroughly studied along with the Part of Speech (grammatical class) of each word. Results across different languages will also be compared in an attempt to discover universally preserved structural patterns
across syntax networks.

\pagebreak
\tableofcontents
\thispagestyle{empty}
 
\listoffigures
 
\listoftables

\pagebreak


\pagebreak

\section{Introduction}

The origin of languages and the relationship among them, along with the existence of universal grammar rules are far from being a closed matter up to this day. Several theories have arisen to explain how humans went from noises and interjections to syntactically complex languages \cite{web:wiki:originslang}. Notwithstanding the hypothesis, all languages currently spoken come from a series of primitive languages called proto-languages \cite{web:wiki:proto}. The list of Proto-languages has been reduced to nearly 50 ``mother'' languages, whose descendants are believed to share some characteristics. 

Common ancestors from languages have been traditionally determined through comparative linguistics, which has proven to be successful since Franz Bopp provided proof of common ancestry for Latin, Greek and Sanskrit in 1816 \cite{web:wiki:compling}. These methods intend to highlight and understand systematic phonological and semantic similarities between languages. Little attention whatsoever has been paid to the only thing that separates human language from those primitive sounds our ancestors made: syntax.

Syntax connects words to each other in very specific ways. Two words are syntactically connected if they depend
directly on each-other. This means that the words directly adapt to (e.g. by being inflected into a specific gender,number, case, etc.) or complement (e.g. by providing direct additional information that adds nuance to) each other-—usually both aspects are involved to different degrees.

Syntactic connections usually happen within a sentence. These connections are strong beyond how far apart words
might appear in the sentence. As words within a sentence start connecting to each-other, a syntactic tree emerges.
This tree is usually directed, as one of two connected words is designed as “head”, commands the other, and occupies
a place higher up in the hierarchy of the tree, as seen in figure \ref{fig:netpluscorp}a

If we gather syntactic trees from different sentences, words arrange themselves in more complex graphs that might
include loops, as displayed in figure \ref{fig:netpluscorp}b. The role of words as heads might change from one sentence to another, thus complicating analyses that
take directedness into account. Following earlier literature, we will ignore the directedness of syntax in what follows.
For each given language, we will consider a corpus that consists of a large number of sentences in that tongue. We will
consider unique words of that language as nodes, and each two words will be connected if they depend syntactically
on one-another in at least one sentence. This yields a complex network or graph that summarizes typical syntactic
relationships in each language.

Complex networks are mathematical objects that have been used to capture relationships within parts of complex
systems. Their study flourished during the last decades, as they offered straightforwards techniques to describe, create,
and control systems that would otherwise be very difficult to comprehend. Syntax networks have been the object of
earlier study [1–6].

Some of these analyses produce summaries of network properties across languages \cite{andreu2009some, arsiwalla2017morphospace, berwick2016only, cech:rolesyntax}, highlighting how
syntax networks are similar in some senses to other complex networks, and how they differ in important aspects. For
example, as the graphs of other complex systems, syntax networks tend to present a heavy-tailed structure, meaning
that most words present relatively little connections, while a few nodes establish a large amount of links. The later
tend to occupy central roles in the network, keeping it well connected and enabling short average distances between
arbitrary words. This is what is know as a small world network \cite{smallworldStrogatz}, where the distance between two nodes $L$ grows proportional to the natural logarithm of nodes in the network $N$.

A way in which syntax networks differ from other typical graphs is that the former are disassortative
while the latter tend to be assortative. In assortative networks, like social networks, well-connected nodes tend to be connected to others
with abundant connections, sparse nodes tend to be connected to nodes with few connections. In syntax networks,
well-connected words do not particularly tend to connect with each other.

An interesting research line has studied how syntax networks form and mature as speakers acquire full-fledged
language \cite{avena2015network, chomsky2008phases}. In these studies we see how syntax networks undergo a seemingly abrupt transition from a tree-like
structure to a topology with multiple loops, dominated by a densely connected core of salient words. At the transition
point, we can observe fast variations of key graph properties such as their clustering coefficient. We can also observe
how the central role of certain words becomes established.

We expand on some of these earlier work sin syntax networks \cite{andreu2009some, arsiwalla2017morphospace, berwick2016only, cech:rolesyntax} by introducing two novelties: On the one hand, we consider a wider and
more homogeneous set of corpora than before. Earlier analyses considered relatively few tongues, and the corpora of
sentences for each language (used to build the syntax networks) was not always homogeneous from a methodological
point of view. We use the Universal Dependencies Database, which contains sytanctically annotated corpora of nearly 100 languages\cite{web:ud:home}. Our analysis
includes 50 different languages (both modern and death) whose corpora has been annotated with roughly similar
methodologies \cite{web:ud:guidelines}. Each of the corpora included at least 30000 words, which contained a large number of unique words. To further homogenize our data, we restricted our analysis to the 500
most common words (which connections are better sampled than those of sparse nodes).

On the other hand, we use a novel network analysis that encompasses and expands earlier approaches \cite{seoane2022tcs}. We measure
a larger set of network properties, and use them to define a morphospace of nodes (i.e. of words) within the network.
Principal components of this morphospace define the dimensions across which nodes present a larger variability,
thus defining the most salient sets of different words. This methodology is used to cluster nodes into topological
communities-—i.e. classes of words that have similar topological properties (and, hence, role within the syntax graph). Our analysis enables a powerful comparative study of syntax networks across all the available languages.

This work is structured as follows: In the Methods section we discuss the procedure that has been followed to create syntax networks. This includes some pre-processing that was necessary to filter out languages with low-quality corpora. We also introduce two analyses that will be performed, and that constitutes the results reported in the Results section: A first analysis consists of a comparative study based on global topological properties measured for the complete syntax network of each language. This approach intends to re-create earlier works \cite{andreu2009some, arsiwalla2017morphospace, berwick2016only, cech:rolesyntax}, and update them with better datasets and network characterization. We reveal a loose phylogeny of languages based on syntactic properties, but which is plagued by misplaced tongues that intrude and invalidate this approach. We thus explain away odd associations that have been observed in the literature, but which have never been properly questioned (e.g., Chinese and English have been reported as `similar' \cite{liuclusters, cong:approach}). A second analysis is performed in a network-by-network manner, and allows us to derive the overall scaffolding of each syntax network. This relies on the measurement of topological properties for individual nodes (i.e., words), and a subsequent clustering and classification of words according to their topological role within their network. We have performed this analysis for all $50$ tongues in our dataset. However, we focus on Spanish to illustrate how this methodology sheds light on network structure. We have uncovered a shared universal structure across all languages studied, of which Spanish is a good representative. This structure is discussed and characterized. We briefly showcase a few tongues that present interesting features (while still falling squarely within the overall, universal shape). Finally, in the Discussion section we summarize our results and contextualize them within the existing literature. We also speculate about the implication of our results for comparative linguistics based on syntax networks and for neurolinguistics.

\pagebreak
\section{Methods}

\subsection{Creating Syntax Networks from Annotated Corpora}

\subsubsection{Syntax networks}
\begin{figure}
    \centering
    \includegraphics[scale = 1.14]{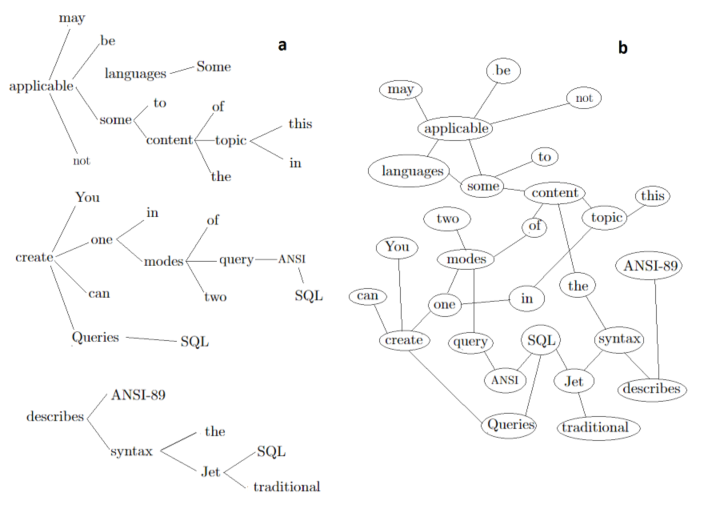}
    \caption{\textbf{Syntactic dependencies of the following text: ``Some of the content in this topic may not be
applicable to some languages. You can create SQL queries in one of two ANSI SQL
query modes: ANSI-89 describes the traditional Jet SQL syntax''. a} Syntactic trees. \textbf{b} Syntactic network }
    \label{fig:netpluscorp}
\end{figure}

Words in a sentence establish relationships of dependency based on rules that vary from language to language. Linking these words hierarchically based on their dependence yields syntactic trees. This tree is usually directed, as one of the words forming the link acts as “head”, pointing to the other, and occupying a higher place in the hierarchy of the tree, as seen in figure \ref{fig:netpluscorp}a. Collecting trees from different sentences and putting them together results in more complex graphs, as the one displayed in figure \ref{fig:netpluscorp}b. Directedness present in the syntactic trees will be ignored in our syntactic networks, following earlier literature \cite{andreu2009some, arsiwalla2017morphospace, berwick2016only, cech:rolesyntax}

More rigorously, a syntax network consists of nodes made up of individual word/token. A token can be a word, a digit, a punctuation mark or a symbol. The nodes constitute a collection $V$ of individual tokens $v_i \in V$ with $i=1, \dots N_t$ and $N_t$ the number of individual tokens. The edges constitute yet another set $E = \{e_k, =1, \dots, N_e\}$ with $N_e$ the total number of couple of tokens that are connected. Each edge consists of a couple of connected nodes, such that $E_k = (v_i, v_j)$ for some words $v_i$ and $v_j$ that are syntactically connected in some sentence. We can codify this information in a square adjacency matrix of size $N_t$, $A$, such that $a_{ij} \in A$ and $a_{ij} = 1$ if and only if $e_{ij} \in E$, and $a_{ij} = 0$ otherwise. Syntax relationships are directed (one word commands the other in a syntactic relationship). However, given that we have ignored directness. $a_{ij} = a_{ji}$, making our matrix symmetric.

\subsubsection{Data Source}

The data used in this study comes from the Universal Dependencies project \cite{web:ud:home}. According to their own website, it is ``a framework for consistent annotation of grammar (parts of speech, morphological features, and syntactic dependencies) across different human languages''. Universal Dependencies (UD) is an open community with over 300 contributors that have produced more than 200 treebanks in 100 languages, with many more on the way. A treebank is basically an annotated(as discussed below) corpus of sentences from some source (e.g. a book or collection of books, newspapers, etc.).

Even if UD is an open community, there exist certain guidelines that ensure the quality and homogeneity of its contributions \cite{web:ud:guidelines}. This has made UD a paramount source of data for other related areas of research such as Natural Language Processing\cite{article:wakawaka}.

\begin{figure}
    \centering
    \includegraphics[scale=0.8]{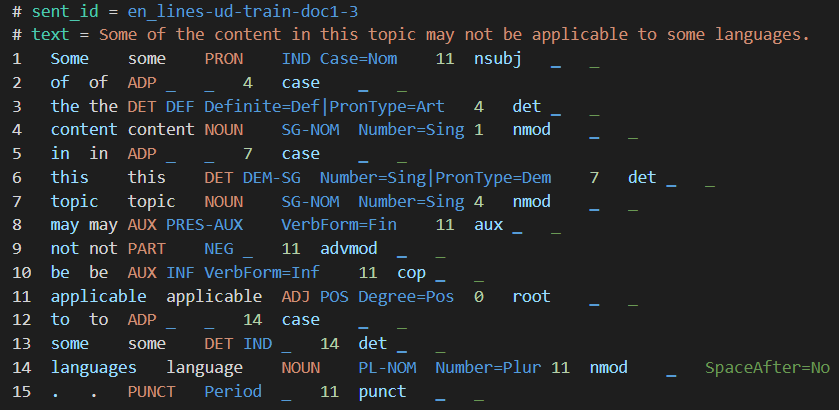}
    \caption{\textbf{Screenshot from corpus with an example of the annotated sentence ``Some of the content in this topic may not be applicable to some languages''}}
    \label{fig:annotated_corpus}
\end{figure}

The whole dataset consists of 100 languages leading to over 200 annotated corpora. Some languages have just one corpus--others, many. Each corpus consists of a collection of annotated sentences, as in Fig. \ref{fig:annotated_corpus}. This annotation allows us to recover, for each individual word within a sentence, its grammatical (or Part Of Speech, POS) category, its lemmatized form (e.g. "are" and "were" are inflected forms of the lemmatized verb "be"), and what other word within the sentence commands it (i.e. to which one each word is syntactically connected). This information allows us to reconstruct syntactic trees for individual sentences (as in Fig. \ref{fig:netpluscorp}a) and, by merging all sentences in a corpus, a complete syntax network (as in Fig. \ref{fig:netpluscorp}b)."  

\begin{table}
\centering
\begin{tabular}{|c|c|c|c|c|}
\hline
\textbf{POS tag} & \textbf{Name} & \textbf{Class} & \textbf{Function} & \textbf{Example}  \\ \hline
ADJ & Adjective & Open class & Typically modify nouns & green    \\ \hline
ADP & Adposition & Closed Class & Prepositions and postpositions & in    \\ \hline
ADV & Adverb & Open class & Typically modify verbs  & well    \\ \hline
AUX & Auxiliary & Closed Class & Accompanies the lexical verb & should    \\ \hline
CCONJ & Coordinating conjunction & Closed Class & links words w/o subordination & and    \\ \hline
DET & Determiner & Closed Class & Determine nouns & the    \\ \hline
INTJ & Interjection & Open class & Exclamation purposes & ouch    \\ \hline
NOUN & Noun & Open class & Denote things or ideas & boat    \\ \hline
NUM & Numeral & Closed Class & Expresses quantity & 3    \\ \hline
PART & Particle & Closed Class & Encode info for other words & not    \\ \hline
PRON & Pronoun & Closed Class & Substitute nouns & you    \\ \hline
PROPN & Proper noun & Open class & Specific nouns & Mary    \\ \hline
PUNCT & Punctuation & Other & Delimit linguistic units & ,    \\ \hline
SCONJ & Subordinating conjunction & Closed Class & subordinate words/constituents & XPOS    \\ \hline
SYMB & Symbol & Other & Word-like, but not a word & @    \\ \hline
VERB & Verb & Open class & Signal events and actions & run    \\ \hline
X & Other & Other & Non-classifiable mostly & dfhjo    \\ \hline

\end{tabular}
\caption{\textbf{Universal part of speech tags. Description and examples of each of them \cite{web:ud:upos}}} 
\label{tab:upos}
\end{table}

Note that not every language showcases all of these parts of speech, and criteria to classify each word to a given POS may differ between databases.

\begin{figure}
    \centering
    \includegraphics[scale = 0.8]{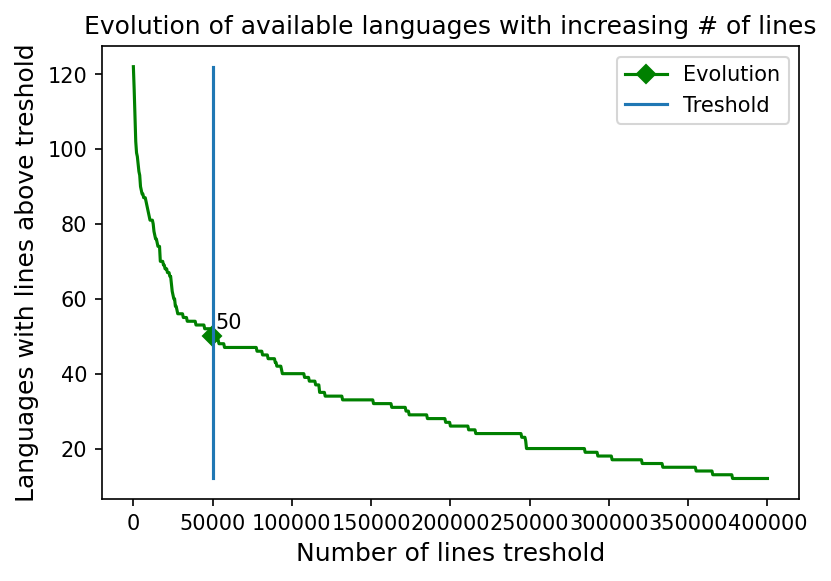}
    \caption{\textbf{Num. lines vs. Num. languages}. Figure shows number of languages in our database that have a corpus with more than a certain number of lines }
    \label{fig:lineslangs}
\end{figure}

Some previous studies compared syntax networks across languages using corpora of different sizes and with different criteria to establish dependency. The UD offers a chance to compare more consistently across languages. From all languages available, we decided to use only those that had at least $N_L = 50000$ lines in their ConLL-U files. This basically ensured a minimum of 30000 tokens present in sentences across their corpora. Fig.\ref{fig:lineslangs} shows how many languages we could have kept if we had taken different cutoff $N_L$. With our threshold, we are left with 50 languages. Table \ref{tab:langstats} shows how many tokens were read for each tongue.

\subsubsection{Building Networks}

Using all tokens in the sampled 50000 lines would result in a huge syntax network that would include a large amount of odd words that appear only once. To circumvent this last problem, we built the network only for the top $500$ most common tokens--not words-- of each tongue. A token might be a word, but can also be a number, a symbol(\$) or a punctuation mark(.). Using much larger networks would quickly become intractable from a  computational stand---as numerical calculations would take over months. It is also worth to mention that at this point we have excluded those words without syntactical meaning, namely, SYMBOL, X, INTJ and PUNCT POS tags.  Table \ref{tab:upos} shows a summary of all POS tags (i.e. grammatical categories).

In order to handle each word correctly and avoid homographs, words were uniquely identified by both its word FORM and POS tag. This way, the noun``well'' can be differentiated from the homographic adverb.

Additionally, for convenience, we run our network analysis only on the largest connected component. For some of the studied languages, this results in networks with slightly less than $500$ nodes (words). Table \ref{tab:langstats} summarizes the eventual size of the largest connected component for each tongue


\begin{table}[!ht]
    \centering
    \begin{tabular}{|l|l|l|l|l|l|}
    \hline
        \textbf{Language} & \textbf{Nodes GCC inflect} & \textbf{Edges  inflect} & \textbf{Nodes GCC lemma} & \textbf{Edges  lemma} & \textbf{Tokens} \\ \hline
        Ancient Greek & 488 & 2638 & 500 & 8283 & 39279 \\ \hline
        Arabic & 494 & 2890 & 499 & 5284 & 39185 \\ \hline
        Armenian & 482 & 2372 & 500 & 5811 & 41252 \\ \hline
        Basque & 494 & 3602 & 500 & 6082 & 39892 \\ \hline
        Belarusian & 472 & 1662 & 496 & 3762 & 36028 \\ \hline
        Bulgarian & 489 & 2694 & 498 & 5299 & 40158 \\ \hline
        Catalan & 494 & 3349 & 500 & 5118 & 43775 \\ \hline
        Chinese & 500 & 3644 & 500 & 3637 & 44608 \\ \hline
        Classical Chinese & 500 & 7050 & 500 & 7055 & 30611 \\ \hline
        Croatian & 491 & 2339 & 500 & 5491 & 43919 \\ \hline
        Czech & 486 & 1970 & 500 & 4890 & 38614 \\ \hline
        Danish & 492 & 4069 & 497 & 5973 & 42891 \\ \hline
        Dutch & 493 & 3666 & 495 & 5061 & 39952 \\ \hline
        English & 496 & 4241 & 499 & 5751 & 42486 \\ \hline
        Estonian & 470 & 2026 & 499 & 5446 & 40653 \\ \hline
        Finnish & 480 & 1892 & 500 & 6197 & 41102 \\ \hline
        French & 490 & 3090 & 498 & 4905 & 43499 \\ \hline
        Galician & 499 & 3965 & 500 & 6422 & 42435 \\ \hline
        German & 496 & 3291 & 500 & 6012 & 42349 \\ \hline
        Hebrew & 494 & 3281 & 500 & 5476 & 37306 \\ \hline
        Hindi & 495 & 4884 & 499 & 5904 & 43264 \\ \hline
        Icelandic & 500 & 5168 & 500 & 9884 & 43447 \\ \hline
        Indonesian & 498 & 3764 & 499 & 4897 & 41518 \\ \hline
        Irish & 496 & 4532 & 500 & 6510 & 44046 \\ \hline
        Italian & 491 & 2935 & 499 & 5040 & 41603 \\ \hline
        Japanese & 499 & 4133 & 500 & 4508 & 42686 \\ \hline
        Korean & 457 & 1483 & 456 & 1503 & 39367 \\ \hline
        Latin & 497 & 6227 & 498 & 9266 & 39752 \\ \hline
        Latvian & 483 & 1986 & 500 & 5509 & 39721 \\ \hline
        Lithuanian & 489 & 1927 & 500 & 5854 & 41720 \\ \hline
        Naija & 499 & 6342 & 499 & 6466 & 32722 \\ \hline
        Norwegian & 495 & 4131 & 500 & 6208 & 41850 \\ \hline
        Old Church Slavonic & 494 & 3297 & 499 & 7785 & 34768 \\ \hline
        Old East Slavic & 491 & 2446 & 498 & 6493 & 32470 \\ \hline
        Old French & 497 & 4707 & 500 & 4645 & 38552 \\ \hline
        Persian & 499 & 4808 & 499 & 6222 & 43028 \\ \hline
        Polish & 482 & 2561 & 498 & 4239 & 35445 \\ \hline
        Portuguese & 489 & 3018 & 499 & 5785 & 42305 \\ \hline
        Romanian & 499 & 4615 & 500 & 8042 & 41482 \\ \hline
        Russian & 483 & 1985 & 499 & 4987 & 42289 \\ \hline
        Scottish Gaelic & 498 & 5258 & 499 & 5759 & 40680 \\ \hline
        Serbian & 491 & 2452 & 499 & 5620 & 43991 \\ \hline
        Slovak & 471 & 1509 & 498 & 3810 & 35708 \\ \hline
        Slovenian & 471 & 1933 & 498 & 4449 & 42368 \\ \hline
        Spanish & 494 & 2852 & 500 & 5318 & 43337 \\ \hline
        Swedish & 493 & 3842 & 499 & 6032 & 40204 \\ \hline
        Turkish & 491 & 2130 & 500 & 5182 & 37678 \\ \hline
        Ukrainian & 466 & 1682 & 498 & 4122 & 39110 \\ \hline
        Urdu & 496 & 5386 & 496 & 5411 & 44957 \\ \hline
        Western Armenian & 474 & 1848 & 499 & 5319 & 39788 \\ \hline
    \end{tabular}
    \caption{\textbf{Language stats for each studied tongue. Results include Giant connected components nodes and edges for both inflected and lemmatized form networks, as well as the original number of tokens present in each corpus}}
    \label{tab:langstats}
\end{table}

\FloatBarrier

\subsection{Data Analysis}
\subsubsection{Language Comparative Analysis}
\label{sec:langcomp}

Building up on earlier work \cite{cech:rolesyntax,cong:approach, liuclusters},we want to test whether the topological structure of syntax networks is preserved over language evolution. To this end, we measure a series of global properties of the each of our 50 languages' syntax networks. Then, we use these properties to cluster together tongues that have similar topological structures. We perform this analysis for both lemmatized and inflected networks. The global properties that we have measured are the averages over all nodes of the node-wise quantities listed on Table \ref{tab:primaryproperties}: 

\begin{table}[!ht]
    \centering
    \begin{tabular}{|l|}
    \hline
        \textbf{Primary properties} \\ \hline
        Degree \\ \hline
        Eigenvector Centrality \\ \hline
        Betweenness Centrality \\ \hline
        Closeness centrality \\ \hline
        Harmonic Centrality \\ \hline
        Pagerank \\ \hline
        Core Number \\ \hline
        Onion layer \\ \hline
        Effective size \\ \hline
        Node clique number \\ \hline
        Number of cliques \\ \hline
        Clustering \\ \hline
        Square clustering \\ \hline
        Constraint \\ \hline
        Component Size \\ \hline
    \end{tabular}
    \caption{\textbf{Primary network properties.} These quantities can be measured for each individual node, and each captures a relevant topological property of the node. For example: (i) clustering measures how many triangles appear in the network, (ii) eigenvector centrality and pagerank convey an idea of how central a node is in the network, (iii) core numbers and onion layer speak about how long a node lasts when you peel off a network by removing the most outward nodes, (iv) effective size or constrain capture whether the neighborhoods of two neighbor nodes are similar, etc. Exact definitions can be found at the networkx Python package documentation\cite{web:nx:docs}. Averaging these properties over all nodes results in global topological qualities that characterize not individual nodes, but networks as a whole.}
    \label{tab:primaryproperties}
\end{table}

 All these properties were computed using \textit{networkx} Python library \cite{web:nx:docs}, and complete definitions may be found in its documentation. Once these networks are represented in an 15-dimensional space, corresponding to our primary properties, they will be brought to a lower dimensional space where these languages may be classified more easily.

\subsubsection{Topological communities in individual syntax networks}

    The analysis described in the previous section is an effort to replicate and expand (with more topological properties and more tongues that are sampled more homogeneously) some earlier works on syntax networks. Opposed to that, in this section we introduce a recently-developed methodology to study complex networks \cite{seoane2022tcs}, which we apply for the first time to the study of syntax. While the analysis in the previous section measured global network properties and comprised a comparison between all languages, our novel method focuses on each individual network separately and intends to reveal salient topological network features that constitute an overarching scaffold of each syntax network. 
    
    We have performed the analysis described in this section for all $50$ tongues in our dataset. For the sake of illustration, this thesis focuses on the case of inflected Spanish alone (while some wider comments are included for other tongues). For the reminder of this section we assume that we work on the network of that language only. A first step in this analysis is to measure as many topological qualities of each individual node as possible. Hence, for each Spanish word, we measure all the so-called primary properties listed in table \ref{tab:primaryproperties}. We enrich this list in two ways. On the one hand, we would like to know whether a node is similar to its connections. Therefore, for each node we measure the average of each primary property over its neighbors. Note, e.g., that if words tend to connect with other words that are similar, a node's primary properties will correlate positively with their averages over neighbors. On the other hand, we would like to know whether nodes are specific---i.e. whether they connect with words of a unique kind, or whether they are `promiscuous' and not very selective. To this end, we compute the standard deviation of each primary property over each node's neighbours.
    
    All these measurements result in a feature vector of $45$ ($15$ primary plus $30$ derived) properties for each node. Which are salient patterns within such quantitative description of the network? If all words have similar values of, e.g., clustering, then clustering does not reveal differences between sets of nodes within the network. On the other hand, if, say, core numbers differ greatly from some nodes to others, then this quantity sets different nodes apart from others---and thus reveals collections of words that might play different topological roles that require, in this example, low versus large core numbers. 
    
\begin{figure}
    \centering
    \includegraphics[scale = 1.1]{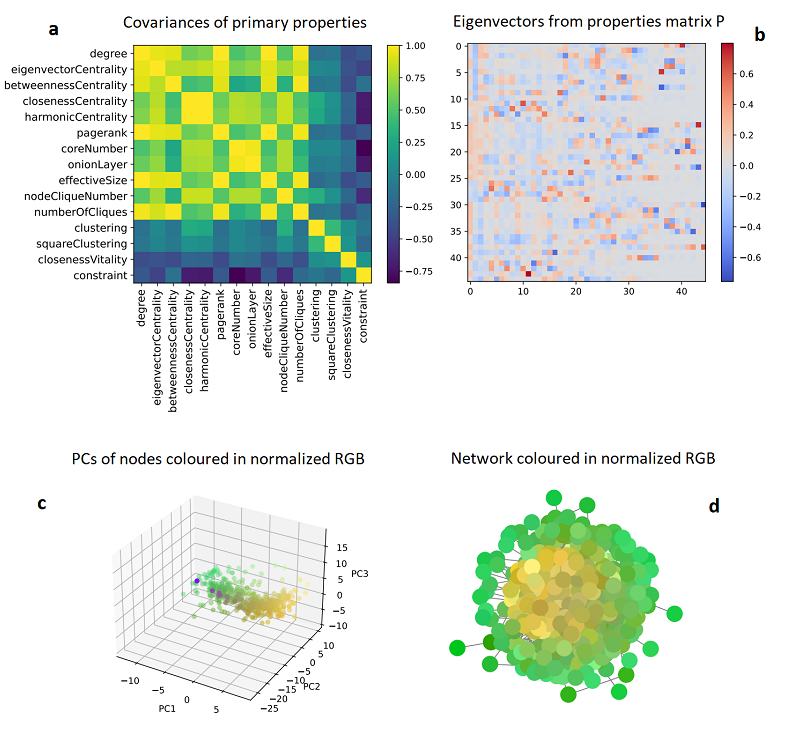}
    \caption{\textbf{Node properties and eigenspace representation summary for inflected Spanish. a} Heat map of covariances of primary node properties, as listed in Table \ref{tab:primaryproperties}. \textbf{b} Heat map of eigenvectors derived from all node properties $P$ as described in section 2.2.2. \textbf{c} Scatter plot of the first three principal components, coloured in normalized RGB colours according to the normalized first three principal components \textbf{d} Network coloured in normalized RGB colours according to the normalized first three principal components. }
    \label{fig:introanal}
\end{figure}

    We use Principal Component Analysis (PCA) to reveal across which dimensions do nodes present the most variety. Fig. \ref{fig:introanal}a shows a typical example of correlations between different node properties, and Fig. \ref{fig:introanal}b shows the eigenvectors of this matrix. Eigenvectors capture the combinations of node features across which we find more variability (with eigenvector importance decreasing left to right). It is not the goal of this thesis to characterize what does each of these eigenvectors mean. However, among the most prominent dimensions of node variability we find a component that separates between more central and more peripheral nodes and also another component that seems to form a gradient of assortativity (in a nutshell, assortative nodes tend to connect to others similar to themselves, while antiassortative nodes tend to link with others specifically dissimilar to themselves). 
    
    The attained Principal Components (PCs) will serve us to variously visualize and classify kinds of nodes found within each individual network. We use the first three PCs to build a ``node morphospace'' (the morphospace for Spanish words is shown in Fig. \ref{fig:introanal}c) that segregates into an abstract space of nodes that are topologically different from each-other. Needless to say, nodes that are similar appear clustered in this volume. Additionally, nodes in Fig. \ref{fig:introanal}c are colored after these first three PCs (red for PC1, green for PC2, and blue for PC3). Projecting these colours into the syntax network layout (Fig. \ref{fig:introanal}d) gives us an idea of whether nodes that are topologically similar are located close to each-other in the network or not. Note that traditional methods that detect communities in complex networks \cite{fortunato} are based on actual proximity of nodes within the network. Our novel method allows us to sort nodes after their topological role, recognizing that they might be similar even if they are not directly connected. 
    
    We perform a clustering study in the PC space of node properties. This groups up nodes that are topologically similar to each other, and allows us to extract ``Topological Communities'' (TCs)---i.e. collections of nodes that play a similar role regarding how the network is structured. We will use these TCs to characterize syntax networks, and extract further information about their structure.

\pagebreak
\section{Results}
\subsection{Comparative study of global topological properties in syntax networks across languages}

\begin{figure}
    \centering
    \includegraphics[scale = 1.05]{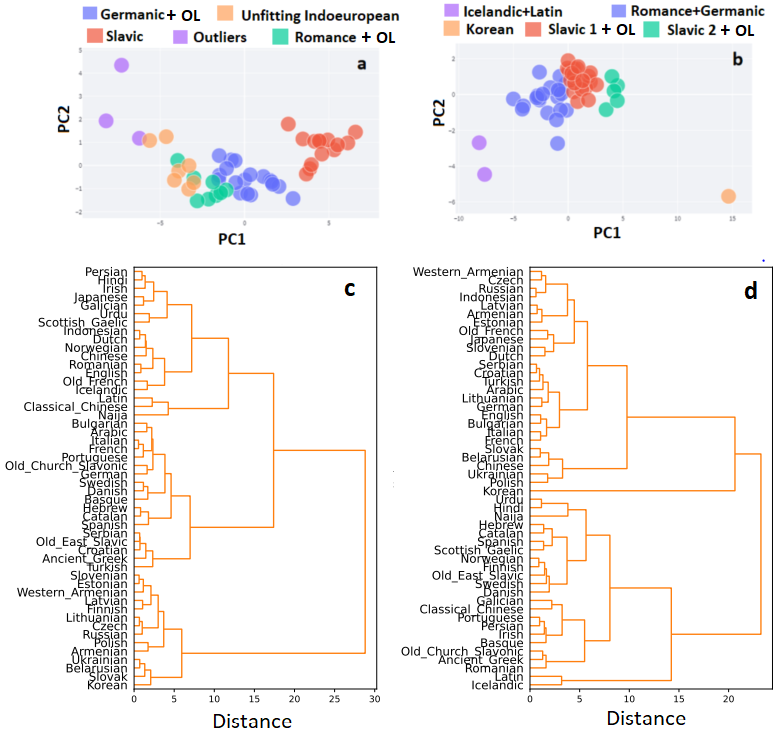}
    \caption{\textbf{Language comparison accross principal components extracted from mean properties as described in section 2.2.1. Note that OL stands for Outliers. a} First two principal components from mean properties of inflected languages coloured by hierarchical clustering communities. \textbf{b} First two principal components from mean properties of lemmatized languages coloured by hierarchical clustering communities. \textbf{c} Dendrogram showing relationships among inflected languages based on hierarchical clustering. \textbf{d} Dendrogram showing relationships among lemmatized languages based on hierarchical clustering}
    \label{fig:langcomp}
\end{figure}

For inflected and lemmatized networks (built as explained in Sec. 2.1) we have computed a series of global topological properties (as explained in Sec. 2.2 and summarized in Table \ref{tab:primaryproperties}). In Fig. 4 we visualize these topological qualities by projecting them into the first principal components (Fig. \ref{fig:langcomp}a for inflected syntax networks, Fig. \ref{fig:langcomp}b for lemmatized syntax graphs). We use these properties to cluster together languages with similar topological properties and create dendrograms (Figs. \ref{fig:langcomp}c and d for inflected and lemmatized respectively)

The clustering of tongues according to its network properties reveal some interesting factors. There are indeed some relevant communities that somehow validate our initial hypothesis. But the name of these communities are purely indicative, as they represent the dominating community within the cluster. Following this trend, we labelled the Germanic+OL and Romance+OL groups according to ancestry of most of their tongues. However, the OL stands for Outliers, indicating that all families are infiltrated by tongues that are distant in evolutionary-relatedness terms. This becomes very patent when we look at the resulting dendrograms, where odd couples such as Japanese and Galician, Indonesian and Dutch, or Hebrew and Catalan are paired together as closest to each-other in the space of topological properties. The takehome message here is that topological properties are not a good guideline to building family trees of human languages. This must be so because these topological properties are not faithfully passed down as new languages evolve from their ancestors---see, e.g., how much Latin differs from its dialects.

Notwithstanding these outliers, there are some interesting features that need to be analysed. For the inflected principal components, it is interesting to see how European languages fall on the chart as if they were trying to mimic the location of their home countries. Romance languages mostly fall on the southwest, with the Germanic on top of them, leaving the Slavic family lying on the east. Note that principal components could be presented with opposite signs, hence inverting this map. Anyways, this could suggest that Germanic tongues are the syntactical bridge joining Romance and Slavic languages. Orange points mainly belong to Indoeuropean languages of Celt and Indo-Iranian and Celtic origins such as Hindi, Urdu, Irish, Scottish or Persian, which is super interesting to say the least. The Outliers are Naija, Classical Chinese and Latin, which have little to do between them.

Lemmatized principal components on the other hand, leave all languages much less spread out, with counted exceptions. The groups are somehow maintained. The now Romance+Germanic group has absorbed languages from both inflected Romance and inflected Germanic groups, which now lie on the rightmost part of the Romance+Germanic cluster. This group has also absorbed most of the inflected tongues from the unfitting Indoeuropean family, which now lie on the leftmost part of the Romance+Germanic cluster. The Slavic languages still seem to have some distinct features that allow for decent grouping, although a bunch of foreign languages have fallen on their group too, like Chinese and Japanese. 

Three outliers stand out. On one hand, Icelandic and Latin; on the other Korean. All of them, specially Korean, are now far from their inflected neighbours. All networks--or almost--change when they are lemmatized. But in the case of Latin, and Icelandic \cite{web:wiki:icelandic}, which are heavily inflected languages with genders, cases and numbers, they seem to have changed too much with respect to the others. Korean apparently represents the opposite trend. This language\cite{web:wiki:korean} has no grammatical gender, rarely specifies number and hardly ever conjugates verbs. This makes its lemmatized network undergo little changes with respect to the inflected one, leaving it far from all other tongues.

Table \ref{tab:meanpropslangs} shows a comparison of mean properties for both types of networks. As expected, most properties grow for lemmatized networks, except for the constraint, that decreases notably. Clustering and the four measures of centrality remain nearly the same. The reason why centrality roughly remains the same even if the average node degree has been almost doubled can be attributed to the small world property of the network, which allows all nodes to be separated by short distances even if they are sparsely connected.

\begin{table}[!ht]
    \centering
    \begin{tabular}{|l|l|l|l|l|l|}
    \hline
        Property & Degree & Eigenvector Cent. & Betweenness Cent. & Closeness Cent. & Harmonic Cent. \\ \hline
        Inflected & 13.48306586 & 0.031577318 & 0.00357038 & 0.377450142 & 196.5970975 \\ \hline
        Lemmatized & 22.4769014 & 0.0331785 & 0.002642008 & 0.441594355 & 231.8388034 \\ \hline
        Property & Pagerank & Core Number & Onion Layer & Effective Size & Node Clique Number \\ \hline
        Inflected & 0.00204352 & 7.270636049 & 19.28583852 & 11.74470963 & 3.55855013 \\ \hline
        Lemmatized & 0.002007638 & 12.0530063 & 28.10675895 & 18.24281177 & 5.093201843 \\ \hline
        Property & Num. of Cliques& Clustering & Square Clustering & Constraint & Neigh. mean Degree\\ \hline
        Inflected & 21.55093173 & 0.14157465 & 0.071411221 & 0.202462651 & 35.39506833 \\ \hline
        Lemmatized & 75.4545371 & 0.223034021 & 0.104057871 & 0.11492954 & 58.85955733 \\ \hline
    \end{tabular}
    \caption{\textbf{Mean properties across languages for inflected and lemmatized networks}}
    \label{tab:meanpropslangs}
\end{table}
\pagebreak
\subsection{Study case of our new methodology: investigating topological communities in the syntax network of inflected Spanish words}

The study in depth of the syntax network derived from inflected Spanish will revolve around the  concept of  Topological community(TC), which is defined as a group of nodes that share similar properties and thus similar roles within the network(as introduced in section 2.2.2).The Part of Speech(POS) tag of each node(presented in section 2.1.2) will also be a relevant factor to be discussed. The relationship of this entity to the TCs will be widely analysed, as well as the distribution of network properties across these same tags and communities.

\subsubsection{Network properties and eigenspace: an overview}

Figure \ref{fig:introanal} shows us the some of the results from the analysis described in section 2.2.2 Normalized first three principal components have been used to create a set of colours in the RGB colour format of the form $(PC_1, PC_2, PC_3)$. Such colours have been used in figures \ref{fig:introanal}c and d.

Glancing at figure \ref{fig:introanal}a, one can observe that trends in covariances are maintained throughout the properties. This means that some variables share similar correlations with each other. This possibly means that a lot of our variables are redundant. By simple observation, three pattern groups can be observed. The first one encloses the degree, eigenvector centrality, betweenness centrality, pagerank effective size and number of cliques; the second one encompasses closeness and harmonic centralities, core number, onion layer and node clique number; and the last one consists of clustering, square clustering, closeness vitality and constraint. Variables in each group are not completely equal, but enclose nearly the same information.

Note that Fig. 5a shows correlations between the primary properties only, while our analysis is based on these plus the $30$ derived properties (as explained in Sec. 2.2.2). They introduce additional correlations and redundancies, as well as adding new information. These redundancies and information contributed are separated when finding out the main PCs (Fig. 5b). Each of these PCs gives us information across which nodes within a network present a wider variability. The information conveyed by each PC is highly non-trivial and requires delicate study beyond the scope of this thesis. These features will be studied elsewhere \cite{seoane2022tcs}. For the current work, PCs are just instrumental to allow us to define TCs (as defined in Sec. 2.2.2), to which we will devote more attention. 

Looking at our Networks, it is easy to appreciate how nodes progressively change their colour, becoming greener the further we get from the core. The green color is represented in normalized RGB format by the tuple (0,1,0), which corresponds to a strong second principal component. Given that the main characteristic of the periphery nodes is their lack of connections, it can be inferred that the second principal component is anticorrelated to the degree of the node.

\subsubsection{Topological communities in the network}

\begin{figure}
    \centering
    \includegraphics[scale = 1.15]{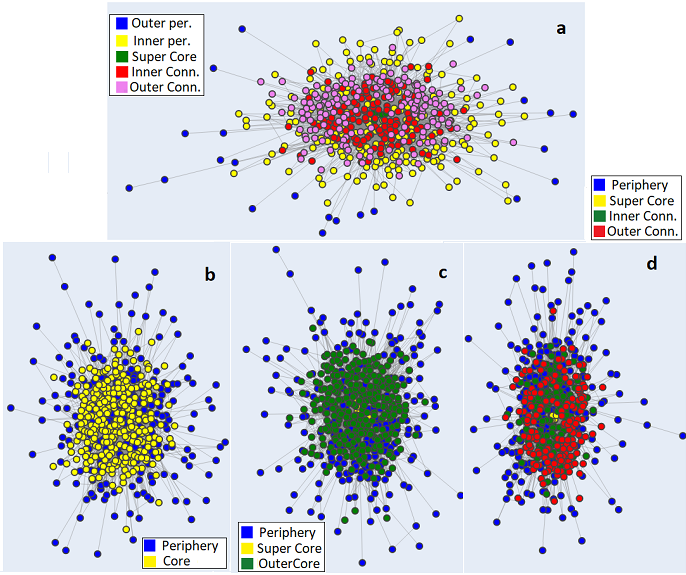}
    \caption{\textbf{Inflected Spanish networks coloured by hierarchical clusters/Topological communities. a} $N_C = 5$. \textbf{b} $N_C = 2$. \textbf{c} $N_C = 3$. \textbf{d} $N_C = 4$}
    \label{fig:TCsines}
\end{figure}

Figure 6 shows the inflected Spanish network colored for fixed numbers $N_C=5$ (Fig. \ref{fig:TCsines}a), $N_C=2$ (Fig. \ref{fig:TCsines}b), $N_C=3$ (Fig. \ref{fig:TCsines}c), and $N_C=4$ (Fig. 6d) of topological clusters. We put the case $N_C=5$ in a prominent place (Fig. \ref{fig:TCsines}a) because we will base much of our analysis on having $5$ topological communities. Starting from $N_C = 2$, one of the communities/clusters splits into two to give birth to a new one. This process is repeated two more times until reaching $N_C = 5$

For $N_C = 2$, there is a clear division that will be maintained for a larger number of communities. There are two groups: A central one with nodes vastly connected, the core; and the periphery, with remote, sparsely connected nodes. Increasing by one the number of topological communities splits up the core, forming a Super Core(SC) or \textbf{Backbone} of just 5 nodes which are very central and an Outer Core that encloses the rest of nodes of the former Core. With  $N_C = 4$, The Outer Core divides into two communities of similar size: the Inner and Outer Connectors(IC and OC). At  $N_C = 5$, the Periphery also splits in two: Inner and Outer Periphery(IP and OP), with most nodes falling on the former.

With regard to the naming chosen for the communities with $N_C$, the reader may have noticed that the difference between inner and outer core and inner and outer periphery is not so big. Nodes in each pair of TCs occupy similar nodes on the network. This naming trend however becomes more patent in most of the syntax networks in other languages.

\begin{figure}
    \centering
    \includegraphics[scale = 0.67]{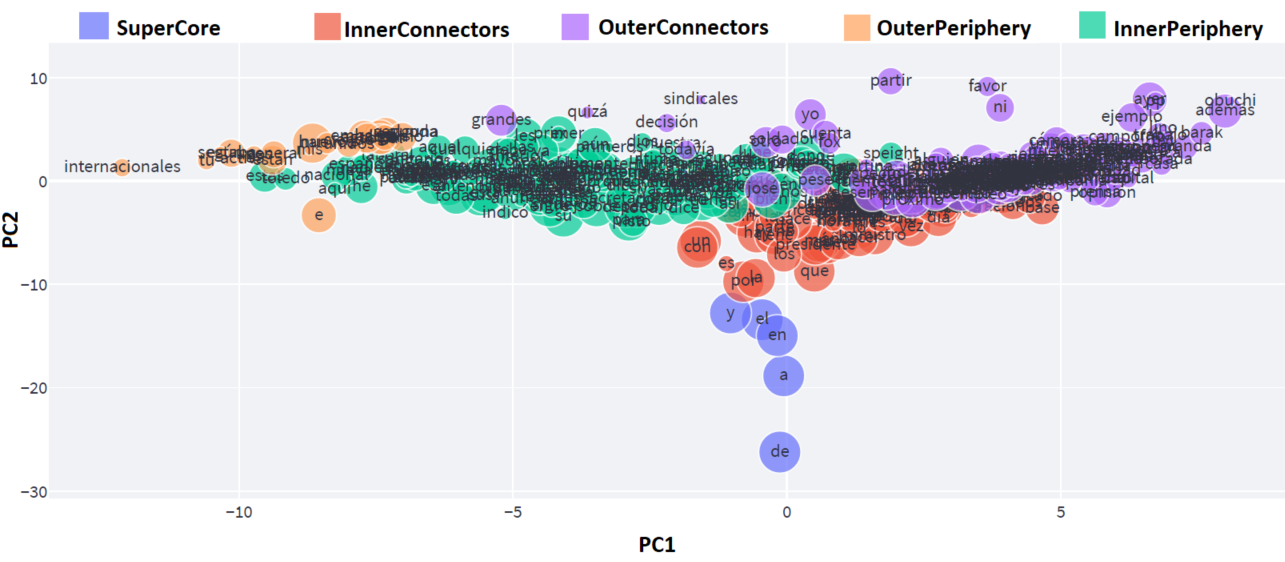}
    \caption{\textbf{First two principal components from all 45 network properties plotted for every node, text indicating the word represented. Each color displays a different topological community as described in section 2.2.2. Node size is proportional to the frequency each node showed up in the corpus.}}
    \label{fig:allwords}
\end{figure}

Figure \ref{fig:allwords} shows every word in the network in the eigenspace of the first two principal components, couloured by TCs for the case $N_C=5$. Words that stand out are mostly those that play a primary role as connectors in the Spanish languages: Adpositions (a, en, de, etc.), determiners (el, la, etc.), conjunctions (que, y, e), etc. Note how abstract these words are, words denoting material objects cannot be found here. All words in the Super Core belong to these group of abstract connectors, but some words from the Inner Connectors may also be found next to them.  Exact composition of each TC by grammatical category can be observed in table \ref{tab:TCscompo}.

Looking at table \ref{tab:TCscompo}, differences between "neighbouring" communities become more apparent. Inner and Outer Connectors have similar compositions, but mostly differ in the amount of proper nouns, much greater in the outer layer. The Inner Periphery is full of adjectives but scarce of nouns, which is the opposite occurring in the Outer Connectors

\begin{table}[!ht]
    \centering
    \begin{tabular}{|l|l|l|l|l|l|l|l|l|l|l|}
    \hline
        \textbf{POS Tag} & \textbf{OP} & \textbf{\%OP} & \textbf{IP} & \textbf{\%IP} & \textbf{SC} & \textbf{\%SC} & \textbf{IC} & \textbf{\%IC} & \textbf{OC} & \textbf{\%OC} \\ \hline
        \textbf{ADJ} & 5 & 0.217391304 & 31 & 0.173184358 & 0 & 0 & 2 & 0.016806723 & 5 & 0.03030303 \\ \hline
        \textbf{ADP} & 1 & 0.043478261 & 10 & 0.055865922 & 3 & 0.6 & 4 & 0.033613445 & 0 & 0 \\ \hline
        \textbf{ADV} & 0 & 0 & 18 & 0.100558659 & 0 & 0 & 4 & 0.033613445 & 13 & 0.078787879 \\ \hline
        \textbf{AUX} & 4 & 0.173913043 & 17 & 0.094972067 & 0 & 0 & 2 & 0.016806723 & 0 & 0 \\ \hline
        \textbf{CCONJ} & 1 & 0.043478261 & 2 & 0.011173184 & 1 & 0.2 & 2 & 0.016806723 & 0 & 0 \\ \hline
        \textbf{DET} & 4 & 0.173913043 & 28 & 0.156424581 & 1 & 0.2 & 1 & 0.008403361 & 1 & 0.006060606 \\ \hline
        \textbf{NOUN} & 1 & 0.043478261 & 13 & 0.072625698 & 0 & 0 & 74 & 0.621848739 & 89 & 0.539393939 \\ \hline
        \textbf{NUM} & 0 & 0 & 6 & 0.033519553 & 0 & 0 & 2 & 0.016806723 & 4 & 0.024242424 \\ \hline
        \textbf{PART} & 0 & 0 & 1 & 0.005586592 & 0 & 0 & 1 & 0.008403361 & 0 & 0 \\ \hline
        \textbf{PRON} & 3 & 0.130434783 & 18 & 0.100558659 & 0 & 0 & 15 & 0.12605042 & 9 & 0.054545455 \\ \hline
        \textbf{PROPN} & 3 & 0.130434783 & 8 & 0.044692737 & 0 & 0 & 2 & 0.016806723 & 43 & 0.260606061 \\ \hline
        \textbf{SCONJ} & 1 & 0.043478261 & 6 & 0.033519553 & 0 & 0 & 0 & 0 & 0 & 0 \\ \hline
        \textbf{VERB} & 0 & 0 & 21 & 0.117318436 & 0 & 0 & 10 & 0.084033613 & 1 & 0.006060606 \\ \hline
        \textbf{Total} & 23 & 1 & 179 & 1 & 5 & 1 & 119 & 1 & 165 & 1 \\ \hline
    \end{tabular}
    \caption{\textbf{Composition of each topological community by POS tag}}
    \label{tab:TCscompo}
\end{table}
\subsubsection{Topological properties across grammatical classes and across topological communities}

Following the previous analysis, we have words in our syntax networks classified according to two different criteria. On the one hand, each word belongs to a grammatical class (nouns, verbs, adjectives, adpositions, etc.). On the other hand, each word belongs to a topological community (backbone, periphery, etc.). We observe that different grammatical classes appear mixed within each TC (a point that we discuss better in the next subsection). TCs have been derived by clustering nodes with similar topological properties, so we expect that there will be a variation of these properties from one TC to another. Is the same true about grammatical classes? Can we say, e.g., that nouns have a higher clustering than adpositions, or that verbs have higher centrality than adjectives? To answer these questions we perform one-way ANOVA analyses that check whether words from different grammatical classes sample the space of topological properties in a similar manner. Results are shown in table \ref{tab:pv:pos}

\begin{figure}
    \centering
    \includegraphics[scale=1.44]{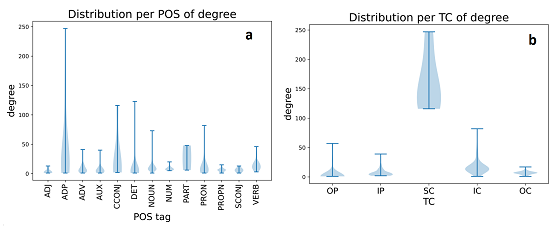}
    \caption{\textbf{Property distribution across different Topological communities and POS tags. a} Distribution of degree across POS tags. \textbf{c} Distribution of degree across TCs. }
    \label{fig:violin}
\end{figure}

 Distributions across POS tags show really small p-values for most properties. This means that in general, different grammatical classes do not sample the topological space similarly.  Figs.\ref{fig:violin}a shows the distribution of degree, to provide some context of how unequal are distributions across grammatical classes.
 
 Distributions across TCs also show that there are indeed major differences in the distributions of all properties. Not only their p-values are inferior to the 0.05 threshold, but most of them are well below $10^{-25}$. The property with the highest p-value, the node degree, has been plotted in figure \ref{fig:violin}b  to illustrate the outraging differences between Topological communities.

\begin{table}[!ht]
    \centering
\begin{tabular}{|l|l|l|}
    \hline
        \textbf{Property} & \textbf{p-value TC} & \textbf{p-value POS} \\ \hline
       degree & 1.31E-130 & 7.16E-17 \\ \hline
        Eigenvector Centrality & 4.42E-121 & 1.43E-14 \\ \hline
        Betweenness Centrality & 1.05E-120 & 2.73E-11 \\ \hline
        Closeness Centrality & 1.02E-113 & 5.91E-34 \\ \hline
        Harmonic Centrality & 5.61E-110 & 4.88E-31 \\ \hline
        Pagerank & 4.63E-132 & 4.81E-17 \\ \hline
        Core Number & 5.93E-45 & 2.86E-19 \\ \hline
        Onion Layer & 2.84E-48 & 1.97E-17 \\ \hline
        Effective Size & 1.71E-133 & 4.58E-18 \\ \hline
        Node Clique Number & 1.07E-59 & 1.50E-20 \\ \hline
        Number Of Cliques & 3.68E-133 & 5.90E-14 \\ \hline
        Clustering & 1.36E-37 & 2.59E-14 \\ \hline
        Square Clustering & 1.09E-27 & 0.000105645 \\ \hline
        Constraint & 4.41E-76 & 3.20E-15 \\ \hline
    \end{tabular}
    \caption{\textbf{p-values for property distributions across POS resulting from conducting one-way ANOVA on our data. The null hypothesis established by this analysis stated that were are no major statistical differences in distributions of properties across each POS tag and TC. p-values indicate the probability that the null hypothesis is true. Values below 0.05 indicate strong probability that the property distribution across these tags and communities are unequal}}
    \label{tab:pv:pos}
    
\end{table}

\subsubsection{Assortativity Analysis}

\begin{figure}
    \centering
    \includegraphics[scale=1.17]{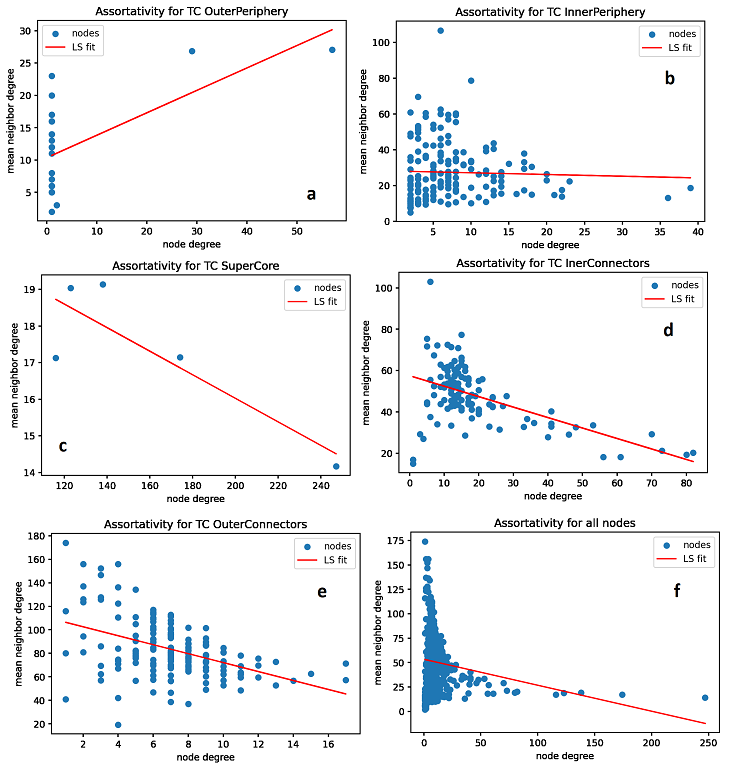}
    \caption{\textbf{Measures of assortativity for the network. a} Scatter plot of Node Degree vs. Mean neighbour degree for the Outer Periphery. \textbf{b} Scatter plot of Node Degree vs. Mean neighbour degree for the Inner Periphery. The red line corresponds to the least squares solution. \textbf{c} Scatter plot of Node Degree vs. Mean neighbour degree for the Super Core/Backbone. The red line corresponds to the least squares solution \textbf{d} Scatter plot of Node Degree vs. Mean neighbour degree for the Inner Connectors. The red line corresponds to the least squares solution. \textbf{e} Scatter plot of Node Degree vs. Mean neighbour degree for the Outer Connectors . The red line corresponds to the least squares solution \textbf{f} Scatter plot of Node Degree vs. Mean neighbour degree for the whole network . The red line corresponds to the least squares solution}
    \label{fig:assort_new}
\end{figure}

Assortativity is a preference for a network's nodes to attach to others that are similar in some way (in terms of their number of connections). For example, in assortative networks, well-connected nodes connect preferably to each other, and sparsely connected nodes connect mostly to themselves as well. A network in which the opposite happens (sparsely connected nodes are rather linked to hubs, while hubs avoid connecting with each-other) is termed antiassortative. Previous studies revealed that syntax networks are antiassortative.  When plotting node degree against that of their neighbours, points in assortative networks should fall close to a line of positive slope. In figure \ref{fig:assort_new} we display a scatter plot of these same variables for 6 sets of nodes: Outer Periphery (Fig. \ref{fig:assort_new}a), Inner Periphery (Fig. \ref{fig:assort_new}b), Super Core (Fig. \ref{fig:assort_new}c), Inner Connectors (Fig. \ref{fig:assort_new}d), Outer Connectors (Fig. \ref{fig:assort_new}e) and all nodes (Fig. \ref{fig:assort_new}f). Each of these plots also shows a line of the form

\begin{equation}
    y = mx + n
\end{equation}

which represents the solution to the least squares problem, where $y$ stands for the mean neighbour degree, $x$ is the node degree, $m$ is the slope of the line and $n$ the intercept. Tables \ref{tab:slopepos} and \ref{tab:sloptc} show the value of such slope $m$ across POS tags and Topological communities.

When analysing the assortativity across POS tags, we see three different trends in the slope of its linear regression: those tags with strongly positive slope, those with a strongly negative one and those of nearly zero slope. Adjectives, Auxiliaries, Numerals and subordinating conjunctions have positive slopes, namely, they are assortative, which is an interesting finding: We know from earlier studies that syntax networks are antiassortative, and here we provide evidence that some specific grammatical classes behave in an opposite (assortative) way. The existence of subset of words that are assortative within syntax networks will be strengthened when looking at TCs. On the other hand, the most disassortative tag is Particle, which in Spanish is just the word ``no''. On the other hand, Nouns, adverbs and pronouns are markedly disassortative. The rest have slopes close to zero. It would be interesting to study more in depth why linking behaviours of nouns and proper nouns are so different.

Looking now at the topological communities, whose assortativity is studied in figure \ref{fig:assort_new}, two communities stand out from the rest: The Super Core, which is highly diassortative; and the Outer Periphery which shows signs of assortativity. Regarding the Super Core, our results make sense since our super core is a a very selected community, with 5 words that join a lot of other words not as well connected. Looking now at the Outer Periphery, it presents positive slope values for its linear regression, which indicates assortativity. In the case of the Outer Periphery this value is remarkably high, $0.384$. Peripheral nodes have few connections, and those they have are usually with other outer nodes with similar properties, which makes their assortativity somehow understandable.

\begin{table}[!ht]
    \centering
    \begin{tabular}{|l|l|l|l|l|l|l|l|l|l|l|l|l|}
    \hline
        ADJ & ADP & ADV & AUX & CCONJ & DET & NOUN & NUM & PART & PRON & PROPN & SCONJ & VERB \\ \hline
        \textbf{2.48} & -0.003 & -1.06 & \textbf{0.344} & -0.033 & 0.001 & -1.416 & \textbf{0.48} & -0.023 & -0.247 & \textbf{0.671} & \textbf{0.451} & 0.03  \\ \hline
    \end{tabular}
     \caption{\textbf{Slope of least square solution of node degree vs. mean neighbour degree across POS tags.} Positive slopes may indicate assortativity }
    \label{tab:slopepos}
    
\end{table}

\begin{table}[!ht]
    \centering
        \begin{tabular}{|l|l|l|l|l|}
    \hline
        Outer Periphery & Inner Periphery  & SuperCore & Inner Connectors & Outer Connectors \\ \hline
        \textbf{0.347} & -0.095 & -0.0321 & -0.506 & -3.81 \\ \hline
    \end{tabular}
    
    \caption{\textbf{Slope of least square solution of node degree vs. mean neighbour degree across topological communities.} Positive slopes may indicate assortativity}
    \label{tab:sloptc}
\end{table}
    
\subsubsection{Studying the composition of TCs in terms of grammatical classes}

    As stated above, we have now two different ways of classifying words within a syntax language. On the one hand, we can sort them according to their grammatical class. On the other hand, we have each word assigned to a TC. Is there a tendency of words from a specific grammatical class to end up in a specific TC? And vice-versa: is there a TC that consists only of words from a given grammatical class? In this section we tackle these questions. To this end, we look at the fractions of POS within a given TC and vice-versa, and we study the generacy of our word groups by looking at entropies, as it follows. 
    
    Let $\{C_i\}$ be the set of POS tags(i.e.\ $C_1=$NOUN, $C_2=$ADJ, etc.). Let $\{T_j\}$ be the set of different TC, with $T_1 \equiv$ Outer Periphery, $T_2\equiv$ Inner Peripher, etc. Given a $T_j$, let us find the fraction $f_{ij}$ of words of each POS class that make it up. Thus: $f_{1j}$ are the fraction of words in $T_j$ that are nouns, $f_{2j}$ are the fraction of words in $T_j$ that are adjectives, and so on. On the other hand, let us find the fraction $p_{ji}$ of words of each POS class that belong to each TC. Thus: $p_{j1}$ are the fraction of nouns that belong to TC $T_j$, $p_{j2}$ are the fraction of adjectives that belong to TC $T_j$, etc. These fractions can be used to define entropies. Thus, the entropy of TC $T_j$ reads:
    \begin{eqnarray}
    H(T_j) &\equiv& -\sum_i f_{ij} \ln \left( f_{ij} \right). 
    \label{eq:entropyH}
    \end{eqnarray}
    Note that the index $i$ runs over different POS classes (nouns, adjectives, etc.). This entropy gives us an idea of how specific a TC is. A low entropy means that a TC only contains words of a POS class---i.e. only words of that class can perform that topological role. A large entropy means that the TC is made up of a well mixed combination of word classes---meaning that any word can implement the topological role of that TC. Similarly, we can define entropies for POS classes: 
    \begin{equation}
    S(C_i) \equiv - \sum_j p_{ji} \ln \left( p_{ji} \right). 
    \label{eq:entropyS}
    \end{equation}
    Note that now the index $j$ runs over the different TCs. If $S(C_i)$ is low, this means that most words in POS class $C_i$ fall in a same TC, thus that they all implement a similar topological role within the syntax network. If $S(C_i)$ is large, this means that words of POS class $C_i$ can belong to any TC, thus that they can play out any topological role in the network.
   
 Let's also define $l_{ij}$ as the number of links connecting words in a POS tag $c_i$ to words in a TC $t_j$. We will call $fl_ij$ to the fraction of $l_{ij}$ over the total number of connections in the graph $N_E$. The same can be done for TC to TC connections. Naming $k_{i_1i_2}$ as the number of links connecting TC $c_{i_1}$ to TC $c_{i_2}$, we can define $fk_{i_1i_2}$ as the fraction of $k_{i_1i_2}$ over the total number of connections in the graph $N_E$.
 
\begin{figure}
    \centering
    \includegraphics[scale = 1.2]{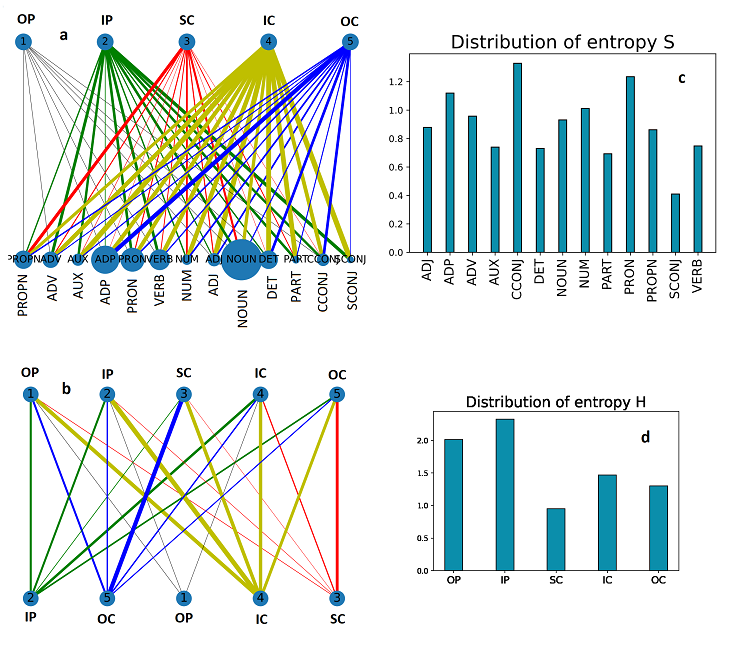}
    \caption{\textbf{Entropy distributions and bipartite networks across POS tags and TCs a} Bipartite network of TCs and POS tag. Bipartite network of TCs and TCs. With of edge joining one community to one POS tag is proportional to the frequency in which this connections happen with respect to all connections in the network.. \textbf{b} Bipartite network of TCs and TCs. With of edge joining one community to another is proportional to the frequency in which this connections happen with respect to all connections in the network. \textbf{c} Breakdown of S entropy distribution across POS tags as described in equation \eqref{eq:entropyS}. \textbf{d} Breakdown of H entropy distribution across POS tags as described in equation \eqref{eq:entropyH}.}
    \label{fig:entrobipar}
\end{figure}

 Figures \ref{fig:entrobipar}c and \ref{fig:entrobipar}d show the breakdown of both S and H entropies across POS tags and topological communities respectively.  Regarding entropy across TC, the super core is as expected the tidiest one--60\% of it are adpositions-- followed by the connectors and the periphery, which seem to be the communities more distributed across tags. This makes sense since the periphery may connect words that do not play a role as a connector, such as nouns or adjectives, and some others that do so but appeared sparsely on our corpus. A good example of this is the word ``e'', which means ``and'' in English and is only used to the detriment of ``y'' when the next word starts with an ``i''. This coordinating conjunction is a natural connector that has landed in the outer periphery most probably due to its low frequency in the sample text. Exact composition of each Topological community can be seen in table \ref{tab:TCscompo}.
  
POS tags leave more room for surprises though. It does not shock us that word classes such as Subordinating conjunctions, which enclose few words with similar roles, have low entropies. It is more appalling however, to see on this end word classes such as Adjectives, Determiner or Verbs.

Figure \ref{fig:entrobipar} also shows bipartite networks displaying the relative frequency with which POS tags and TCs connect to each other, as given by the frequencies $fl_{ij}$ (Fig. \ref{fig:entrobipar}a) and $fk_{i_1i_2}$ (Fig. \ref{fig:entrobipar}a) introduced in this same section. From the POS-Tc bipartite network, it is interesting to note how adpositions, the most important connector, are mostly connected to elements in equal measure. Inner connectors are connected to each grammatical class in nearly equal manner. The TC-TC network is even more revealing. The first thing that catches the eye is probably the disassortativity of the networks. As it was seen before, nodes in our network are not usually linked to nodes with similar properties. This is confirmed by seeing how weak connection between the same TCs are. Outer Periphery, which was shown to be assortative, is the exception.  Its connections with itself are only as weak as connections with other clusters.

\pagebreak
\subsection{Comparison of Topological communities across languages}

In the previous section we have used the Spanish syntax network to illustrate a novel method to analyze complex networks, and how it allows us to extract important insights about syntax. We have performed this analysis on all $50$ tongues with satisfactory sampling in the UD database. This results in a varied collection of results concerning syntax of many languages which will be released in the form of research papers in the future. In this section we discuss a few interesting patterns that we have observed. A prominent one perhaps is that the overarching structure of syntax networks (their decomposition in a core-periphery, as well as the emergence of a selected backbone, and an asymmetric split between non-backbone nodes in the core) seems to be universal across all language networks tested (and likely in languages not included in this study as well). This uniformity of the general structure of syntax networks should not be taken for granted and is not an artifact of the analysis method. Other kinds of networks not discussed here (e.g. networks of collaborations between scientists, connectomes in the brain, networks of flight connections between airports, or networks of political collaboration) result in TC decompositions that differ from the ones found for syntax \cite{seoane2022tcs}. In this section we display briefly some of the cases that differ the most from this universal pattern, and we study similarities for four selected romance languages.

\subsubsection{Network comparison}

\begin{figure}
    \centering
    \includegraphics[scale = 1.31]{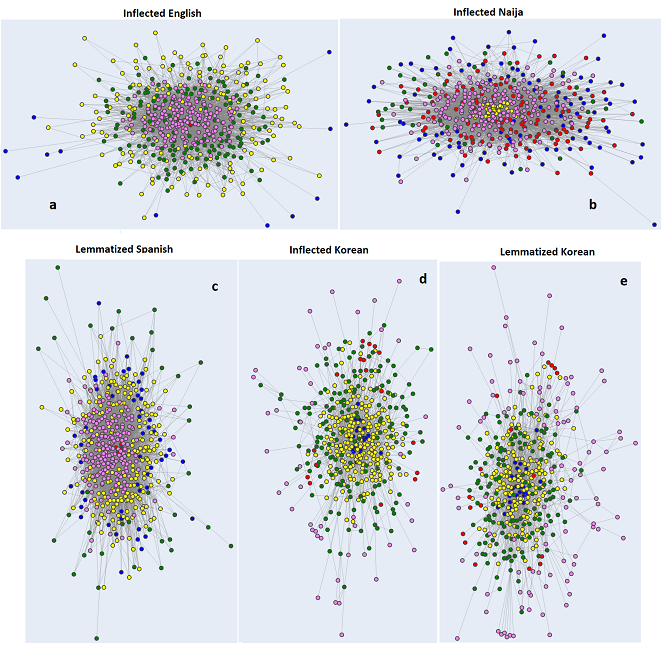}
    \caption{\textbf{Syntactic networks of languages coloured by topological communities derived from hierarchical clustering, $N_c = 5$. a} Inflected English. \textbf{b} Inflected Naija. \textbf{c} Lemmatized Spanish. \textbf{d} Inflected Korean. \textbf{e}} Lemmatized Korean.
    \label{fig:network_compar}
\end{figure}

Syntactic networks have proven to follow a certain structure in most cases. As it was discussed for the case of inflected Spanish, Inflected language Networks usually consist of 5 well distinguished topological communities. In the center, a core made of a scarce backbone surrounded by an abundant layer of inner and outer connectors. These connectors are linked to the most outer-peripheral nodes through the so called inner periphery. This structure is also followed by the inflected English network, which is displayed in figure \ref{fig:network_compar}a. This structure is however defied by some languages, as it is the case of Naija, an English creole language widely spoken in Nigeria. As it could be seen from figure \ref{fig:langcomp}a, Naija is the left-uppermost point, which hinted at some differences in the network. Taking a closer look at the Naija network, portrayed in figure \ref{fig:network_compar}b, it is easy to realize that it does not resemble the network of its mother tongue, English. In this graph, nodes are much better distributed across topological communities. The core is now numerous, and so are the inner connectors, but not as much as in traditional networks. Both peripheral communities and the outer connectors have a similar number of nodes, and the outer-most words look better connected. Besides that, roles of each community in the work do not look so well defined.

It is hard to know the reasons why a creole language could be so syntactically different from its mother. Perhaps variations in the word forms slightly created ``duplicate'' words that allowed for the better distribution of connection responsibilities along the network. Looking at some central nodes in the network we can find words such as ``dis'', or ``dem'', which are just creole written forms of ``this'' and ``them''. But it is likely that the British English forms are still present in the corpus, hence stealing linkages from these central nodes.

Figure \ref{fig:network_compar}c also displays for reference the network of lemmatized Spanish. When comparing it to the network of inflected Spanish, shown in figure \ref{fig:TCsines}a, it is not hard to tell the differences between them. The core basically stays the same, the periphery of the network has shrunk while the connectors have grown massively. These "migrants" are mostly verbs and nouns which were once inflected in the form of tenses, plurals and gendered forms and are now expressed as single nodes with lots of connections between them. This is the case for most inflected languages.

For Korean though, that is not the case. Figures \ref{fig:network_compar}d and \ref{fig:network_compar}e display its inflected and lemmatized networks respectively. As we discussed in section 3.1, Korean grammar allows for inflection, but it is rarely used. This was the reason we wielded to understand why this language was so far apart in figure \ref{fig:langcomp}b. Our hypothesis then was that all lemmatized networks had changed with respect to their inflected forms, while Korean remained the same. Drawing both networks shows that our theories were not that far from reality. Both networks have similar number of nodes in each cluster, although now are the peripheral nodes which get their number slightly increased to the detriment of connectors.

\subsubsection{Iberian Romance Languages Comparison}

Looking at figure \ref{fig:allwords} it is hard not to tell how different are Spanish's main connectors(de, y, a, en, la...) to the vast majority of words. This had us wondering about the possible existence of a universally preserved structural pattern across syntax networks. Would other languages' main connectors show such distinctive features?

\begin{figure}
    \centering
    \includegraphics[scale = 1.12]{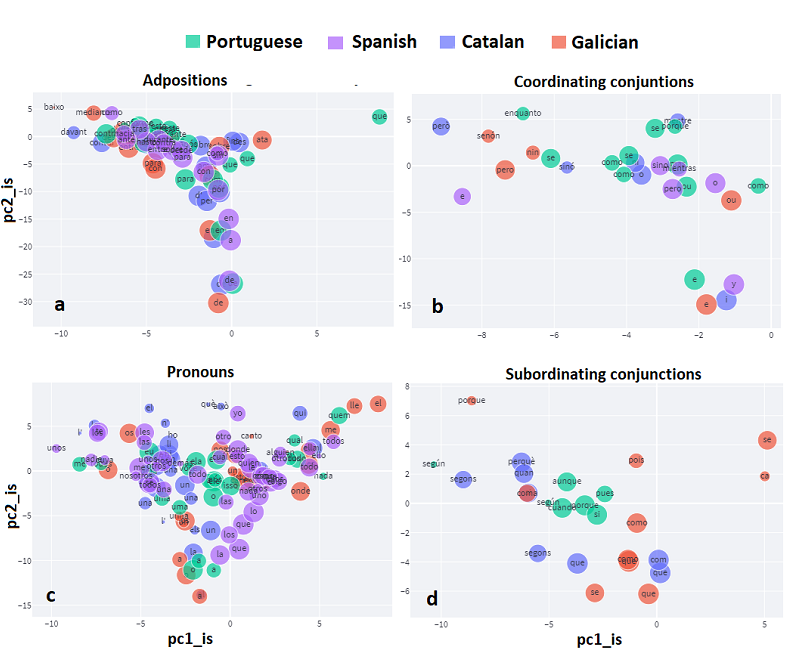}
    \caption{\textbf{Scatter plots of Nodes of Romance Iberian languages projected onto the first two principal components of the Inflected Spanish eigenspace across selected POS tags. a} Adpositions. \textbf{b} Coordinating conjunctions. \textbf{c} Pronouns. \textbf{d} Subordinating conjunctions }
    \label{fig:romibcomp}
\end{figure}

In order to shed some light on this matter, a simple experiment was carried out. Words from selected POS tags were plotted in the same eigenspace for 4 languages that share many similarities: Castillian Spanish, Galician, Catalan and Portuguese. Results from this test can be seen in figure \ref{fig:romibcomp}.

Results in this regard are, to say the least, astounding. In the case of adpositions, displayed on figure \ref{fig:romibcomp}a, there are two groups that stand out from the rest. One is formed by the four ``de'' words, which means of, since and from in all four languages. Their coordinates in the PC space are nearly the same. The other is formed by the words ``em'', ``en'' and ``a'' in Spanish and Catalan. These words can either mean in or to, and their form varies in the last two languages by the influence of French, but their syntactic roles are the same.

Pronouns, portrayed on fig. \ref{fig:romibcomp}b also show a similar cluster for the masculine and feminine personal pronouns ``la'', ``lo'', ``a'', ``o''... For the coordinating conjunctions, shown in figure \ref{fig:romibcomp}c, there is a cluster formed by ``e'', ``i'' and ``y'', which all mean ``and''. These similarities across languages are also found in subordinating conjunctions(fig. \ref{fig:romibcomp}d): ``como'' and ``com'', which could be translated to ``as''; or ``que'', which translates to ``that''. Further analyses of this alleged universality should be carried out in the future with a wider range of words.

\pagebreak
\section{Discussion}

    In this work we take a fresh look at some old scientific questions around syntax networks. Namely: What is their general structure? Is it singular for each individual language, or do we have a universal scaffold repeated throughout? Can we use topological properties of syntax networks to find how different tongues are related to one-another? We tackle these questions by updating the existing literature in two different ways. On the one hand, we use a larger database that allows us to compare more different languages. This database is also more consistent (methodologically) than others used earlier, and its exuberance allows us to study only a subset of tongues (still larger than those of previous studies) for which we can guarantee that all words in our syntax networks have been sufficiently sampled. On the other hand, while we repeat an earlier analysis to check for consistency, we also deploy a novel computational framework for the study of complex networks. 
    
    The earlier analysis that we reproduce consists of measuring global topological properties of syntax networks for different languages, and to use distances in this abstract space to elaborate dendrograms that would tell us how  tongues are related to each other. A hope of this line of research is that evolutionarily related dialects (e.g.\ all those coming from Latin) would share some topological features that also differentiate them from other families (e.g.\ Semitic or Germanic). If this were the case, our derived dendrograms should resemble known linguistic phylogenies. Earlier works have reported findings along these lines, but also odd associations (e.g.\ English and Chinese would often be found close by), that we think have not been properly interrogated \cite{cong:approach,liuclusters}. 
    
    Our results are rather negative in this regard. While we can see some faint clustering of languages from same families, we find that unrelated tongues often intrude branches of the phylogeny to result in unrealistic associations (e.g.\ Japanese and Galician, Indonesian and Dutch, and Hebrew and Catalan appear as couples of closest neighbors). We thus rather conclude that syntax network properties are not good features to elaborate linguistic phylogenies. A reason for this would be that topological properties of syntax networks are not preserved enough through the evolutionary process that yields new dialects. Thus, the fact that a mother tongue has a specific quality  does not imply that its descendants would share that same feature, as it was seen with English and Naija. An alternative explanation (not necessarily incompatible with the previous one) is that the space of possible syntax networks is actually rather small, hence emerging languages will explore areas of this space that have already been sampled by some other tongue, even by far-away ones---thus messing up possible phylogenies. If it is the case that the possible space of syntax networks is thus constrained, we should then observe some kind of regularity, or even universality, in the overall structure across languages. 
    
    The results of our second analysis strongly point to this direction. We have applied a novel method that studies complex networks by measuring node-wise features, and elaborating a {\em morphospace} \cite{raup1966geometric, avena2015network, corominas2013origins, goni2013exploring, seoane2018morphospace, sole2022evolution, arsiwalla2017morphospace} of nodes. This method allows us to distinguish vertices of a graph that are topologically similar to each other, even if they are not connected or even close within the network (unlike traditional community detection methods, which are based on direct connectivity between nodes). Grouping up nodes in this way results in {\em topological communities} that reveal the overarching organization of the network. We find that topological communities across networks are remarkably similar, suggesting a universal organization for syntax networks across languages. While we have implemented this analysis on all $50$ tongues in our data set, both for inflected and lemmatized networks, we focus on Spanish to report on our results. 
    
    We find that syntax networks consist, first, of a dense (and densely connected) core formed by the supercore/backbone and connectors and an abundant (yet sparsely connected) periphery. While words in the core connect both in- and outwards, those in the periphery usually get connected through the core. It is known that syntax networks in children evolve into this configuration from a phase in which they look more tree-like, and in which dense central connections are not quite present \cite{corominas:ontogeny, corominas:chromatic}. Secondly, we observe that a backbone/supercore stands out from within the core. This most central part of the network is diverse (consisting of words from different grammatical classes), it plays a huge role in keeping the network together, and is made up of the most common words of each language. While we still need to look deeper into the details, we seem to appreciate that most words in the syntactic backbone/supercore are notably similar across languages---or, rather, they play similar roles, such as being the most common pronoun, or acting as auxiliary or modal verbs. The backbone is embedded in the larger, densely connected core. Regarding all other core words that do not belong to the backbone, we appreciate consistently across languages (while we report only for Spanish) that they split into two asymmetric topological communities. This divide is very salient, but it has been difficult to determine what exact features lead to it---this falls beyond the early exploratory work presented in this thesis. A patter that we seem to observe is that one of these subgroups is notably assortative (meaning that nodes tend to connect to others that are similar---e.g.\ highly-connected vertices to other highly-connected ones, and sparsely-connected nodes to also sparsely-connected ones). This is a remarkable possibility, because syntax networks are known to be disassortative (highly-connected words tend to link to sparsely connected ones and vice-versa), thus constituting an important exception among most other complex graphs (which are usually assortative). Finally, we observe that the network periphery of syntax networks also splits between two distinct groups along a similar dimension as the divide in the core. 
    
    When pondering the possibility that there is a universally preserved structural pattern across syntax networks, it is important to understand how topological communities emerge from the novel analysis that we apply. In our methodology, a large set of topological properties are measured for each node. Then, a Principal Component (PC) analysis is used to find out along which features (or combinations of features) we find more variability---as well as what patterns are more consistently recovered. This process is dictated by the nodes' properties within the network alone. In other words, dimensions of variability or topological communities are not imposed from an external guess, but rather emerge out of what turns out to be relevant in each individual network. We did not guess beforehand a core-periphery distribution (however well-known this was from previous studies); neither that a backbone would stick out (and that it would be remarkably similar across tongues); nor that we would observe further, asymmetric splits, perhaps along a line of assortative versus disassortative words. All these salient features were dictated by each language's syntax network structure alone, thus making more outstanding the apparent universality. Using this same methodology in other kinds of networks results in other different sets of overarching patterns. 
    
    This work is an exploratory study, as it applied a very novel methodology to a quite complex kind of graphs. Our results are outstanding, but they also require a sober, in-depth analysis. This will be performed in the near future, and will likely be published in one or several scientific papers, along with the main results presented here. Some of the questions that need to be addressed are: While we guess a role for assortativity, what are the specific features that separate each topological community? Can we quantify the degree to which topological communities are universal (e.g.\ by tracking cognates across languages)? Specifically, can we do this for the syntactic backbone? In this sense: Can we reduce the syntactic backbone to a minimum list of organizational and semantic roles that need to be fulfilled by all languages, and whose fulfilling, indeed, makes words so central to syntax? This is a fascinating possibility. What can we learn from our outliers? Fig.\ \ref{fig:langcomp} shows a few languages for which we find slightly odd configurations---which nevertheless broadly match universal scheme uncovered. What is it that makes these tongues different? 
    
    Finally, we think that it is worth to reflect about our findings in neurolinguistic terms. Our work is based on the external form of language, but this is just the reflection of an internal organization. The underlying questions here are: How is syntax generated by the brain? How is this neural circuitry reflected by the externalized form of language---and, hence, by our syntax networks? Can we constrain what circuitry to expect given the observed, external shape of syntax graphs? Naively, we could guess that different syntactic roles are implemented by dedicated circuits: a verb does what a verb does, an adverb what an adverb does, etc.; and these roles must be wired somehow in our gray matter. However, if we attend to Chomsky's minimalist hypothesis \cite{chomsky2008phases, berwick2016only}, there is only one or two single syntactic mechanisms---namely merge and, perhaps less prominently, nest \cite{andreu2009some}. If this is the case, then the neural representation of each individual word might be endowed with the necessary circuitry to be articulated through these mechanisms. Devoted wiring would not be needed for specific syntactic functions alone. Then, other features might serve to sort out words within the brain. Might their topological characteristics within their syntax network play a role here? In other words, can we find a representation for our syntactic backbone that is different from the encoding of words in the rest of the core and periphery, and that also reflects the found asymmetric split in these topological communities? 
    
\pagebreak

\section{References}

\bibliographystyle{plain}
\bibliography{theBibliography}

\end{document}